\documentclass[sigconf]{acmart}

\usepackage{amssymb}
\usepackage{amsmath}
\usepackage{graphicx}

\DeclareMathOperator*{\argmax}{argmax}

\copyrightyear{2018} 
\acmYear{2018} 
\setcopyright{acmlicensed}
\acmConference[GECCO '18]{Genetic and Evolutionary Computation
Conference}{July 15--19, 2018}{Kyoto, Japan}
\acmBooktitle{GECCO '18: Genetic and Evolutionary Computation Conference,
July 15--19, 2018, Kyoto, Japan}
\acmPrice{15.00}
\acmDOI{10.1145/3205455.3205630}
\acmISBN{978-1-4503-5618-3/18/07}

\begin{document}
\title[Inheritance-Based Diversity Measures for Explicit Convergence Control]{Inheritance-Based Diversity Measures\\for Explicit Convergence Control in Evolutionary Algorithms}

\author{Thomas Gabor, Lenz Belzner, Claudia Linnhoff-Popien}
\affiliation{\institution{LMU Munich}}

\renewcommand{\shortauthors}{Gabor et al.}

\begin{abstract}
Diversity is an important factor in evolutionary algorithms to prevent premature convergence towards a single local optimum. In order to maintain diversity throughout the process of evolution, various means exist in literature. We analyze approaches to diversity that (a) have an explicit and quantifiable influence on fitness at the individual level and (b) require no (or very little) additional domain knowledge such as domain-specific distance functions. We also introduce the concept of genealogical diversity in a broader study. We show that employing these approaches can help evolutionary algorithms for global optimization in many cases.
\end{abstract}

%
%



\maketitle

\section{Introduction}

Diversity has traditionally been known as key asset for an evolutionary process. Higher levels of diversity within the population undergoing evolution steer the focus of the evolutionary search towards the exploration of the search space and away from convergence on the already found solutions. This often helps discover better global solutions. The precise handling of the exploration/exploitation dilemma is of central importance for the the success of all importance sampling techniques \cite{spears2013evolutionary,vcrepinvsek2013exploration}.

Thus, many methods have been suggested to observe and subsequently control the level of diversity within a population. These have also been compared in previous studies \cite{squillero2016divergence}. All them introduce a notion of diversity given either (a) as a function of the population or (b) as a relation between an individual and (parts of) the population. This means that we can use diversity measures of type (a) to motivate and evaluate global approaches to increase or decrease the levels of diversity (like an increase of hyper-mutation or migration). Measures of type (b) can estimate the diversity introduced by each single individual and are commonly used as an additional objective to the evolutionary process. Instead of using a classical multi-objective evolutionary algorithm each time we employ individual-based diversity, we can also adjust the selection process to respect diversity in a different manner or use other common techniques to transform multi-objective evolutionary algorithms into a single-objective case.

In this paper, we focus on a specific kind of individual-based diversity: In order to avoid defining (and assessing) domain-specific measures of diversity that need to be adjusted to the specific data types used for the individuals' genomes, we attempt to create a more general approach to diversity that does not directly depend on the structure or contents of the genomes. Instead, we want to leverage information already generated by the evolutionary process in order to give an estimate of the diversity of specific individuals. To this end, we perform a thorough evaluation of the novel notion of \emph{genealogical diversity} \cite{gabor2017genealogical}: We track the genealogical relations between individuals (in an efficient manner) and then assume that closely related individuals are more similar than unrelated individuals regarding the diversity they add towards the population. We show that evolutionary algorithms using genealogical diversity can reach similar levels of performance as those using domain-specific diversity measures and usually achieve better results than other generic approaches to diversity (such as population ensembles).

Our research is originally motivated from a software engineering point of view: Self-adaptation and self-organization are playing an increasingly important role in the design and implementation of large-scale software and cyber-physical systems, the reason being that state-of-the-art methods of optimization are not only able to save effort for human developers but are starting to show the ability to forge solutions that allow for entirely new applications~\cite{de2013software,wirsing2015software}. However, many of the new techniques for autonomous search come with a large amount of parameters that are expensive to fully evaluate. We thus see an inherent benefit from providing means to control the convergence of an evolutionary search process without depending on a specific choice of data structure or search domain.



A short overview of related work is given in the following Section~\ref{sec:relatedwork}. We introduce some diversity techniques in greater detail in Section~\ref{sec:diversity}. We thereby motivate and introduce the approach of genealogical diversity. Section~\ref{sec:results} discusses empirical experiments that justify our approach. Finally, Section~\ref{sec:conclusion} provides a recap on this paper and a glimpse onto future work.

\section{Related Work}
\label{sec:relatedwork}

Diversity is a topic often researched and discussed in literature about evolutionary algorithms. In short, a diverse population features many and by tendency more different genotypes of individuals, which has been known to be a key factor in preventing the problem of \emph{premature convergence}~\cite{eiben2003introduction}. However, oftentimes diversity is only regarded as a tool for an \emph{a posteriori} analysis of the behavior of an evolutionary process \cite{spears2013evolutionary}. Thus, a lot of research has been focused on constructing the evolutionary algorithm in such a way that low diversity is prevented implicitly \emph{by design}. Approaches like island-based models or other spatial structures imposed on the set of individuals can be regarded as designs aiding higher diversity \cite{tomassini2005spatially}. Ensemble methods open up a variety of configurations that can be used by the designer of the evolutionary algorithm to provably increase performance \cite{hart2017constructing}. We deliberately choose a quite simple instance for a comparison in this paper. The full scope of combining the possibilities of diversity-awareness with ensemble learning are up to further research.

The maintenance of population diversity can also be tackled more explicitly: Observing diversity while the evolutionary process is still running allows a watchdog process to intervene whenever it does not fulfill the desired level of diversity \cite{deb2000fast,ursem2002diversity}. In such a setup, diversity is only improved by drastic methods altering the whole course of the evolutionary process in the form of a ``last resort.''

A newer line of research has focused on utilizing the evolutionary process itself to optimize diversity throughout the whole process, i.e., add diversity as a direct objective for the evolutionary algorithm \cite{segura2013using}. This exposes the meta-goal of preventing premature convergence to the evolutionary algorithm and allows engineers to explicitly slow down the convergence process. Naturally, applying a second objective function yields a \emph{multi-objective evolutionary algorithm (MOEA)}, which requires more complex (and thus more computationally expensive) methods of selection \cite{laumanns2002combining}.

A most extensive overview of techniques of adjusting the exploration/exploitation trade-off in evolutionary algorithms can be found in \cite{vcrepinvsek2013exploration}. The authors of \cite{segura2013using} also provide a great overview over various methods to estimate the diversity of a single individual, remarking that distance-based methods have been shown to work best. The described distance functions like the distance of the closest neighbor, however, require consideration of most if not all of the individuals in the population, which comes with substantial computation load for large-scale examples. The authors of \cite{squillero2016divergence} perform an extensive survey of various means to define, measure and augment diversity in evolutionary algorithms. All of these papers also introduce comprehensive taxonomies.

For Particle Swarm Optimization and Differential Evolution approaches the authors of  \cite{vsenkerik2016application} have traced and visualized historic or genealogical relationships of individuals. However, they do not use that knowledge to further steer the evolutionary process.


\section{Diversity in Evolutionary Algorithms}
\label{sec:diversity}

As genetic algorithms maintain a pool of solution candidates at any given point in time, they are a natural fit for an optimization algorithm that searches for multiple local optima at once. However, even whole populations of solution candidates tend to converge to one single local optimum in some scenarios. A remedy discovered in the context of genetic algorithms is the notion of \emph{diversity}: As we shortly discussed in the Introduction, diversity-enhancing techniques exist at the population level or at the individual level. 

In this Section, we introduce our measure of genealogical diversity by deriving it from other measures of diversity we sketch in the course of this Section. First of all, however, we give a short definition of the formal framework we use for evolutionary algorithms.

\subsection{Evolutionary Algorithms}

Let $D$ be the search domain of a given problem we want to optimize. The optimality of a solution candidate $x \in D$ is given via the \emph{objective goal function} $g: D \to \mathbb{R}$. The solution to a maximization problem can thus be written as $\argmax_{x \in D} g(x)$ and likewise for minimization. The objective goal function will usually be the main influence on the fitness of an individual. In general, we define a fitness function $f: D \times \mathcal{P}(D) \to \mathbb{R}$ so that $f(x, P)$ denotes the fitness of the individual $x$ within the population $P$ with $x \in P$. Note that we pass on the whole population to the fitness function so that we can, e.g., respect the diversity of the individual with regard to said population. This also has the immediate effect that the fitness of an individual may vary without any actual change to the genome.

A population is a set of individuals. An individual usually directly represents a solution candidate $x \in D$ so we will use these notions interchangeably. We can thus give the type of a population $P$ as $P \in \mathcal{P}(D)$. In detail, however, an individual is always part of a population and may thus have additional properties like genealogical relations such as, e.g., parents and children. A population is affected by evolutionary operators $o: \mathcal{P}(D) \to \mathcal{P}(D)$. In the evolutionary algorithms described in this paper we use common implementations of recombination, mutation, hypermutation and selection in that order. A series of populations resulting from the iterated application of these evolutionary operators is called an evolutionary process.

The examples in this paper show different instances of an evolutionary algorithm: The most simple one is called \emph{non-diverse} and is directly derived from the setup described so far. Its fitness function $f$ can simply be defined as $$f(x, P) = g(x).$$

Note that when not stated otherwise, for the remainder of this Section we assume to optimize a maximization problem, i.e., maximize the fitness function. All definitions can be trivially adapted to the minimization case.

\subsection{Population-based Diversity}

Population-based methods attempt to increase the diversity within the evolutionary search process without computing a specific diversity value for every single individual. We further discern them into \emph{structural} and \emph{reactive} methods. The latter usually observe some diversity measurement throughout the evolutionary process and employ some methods to increase diversity once a state of little diversity has been observed. Commonly, these measures could be to increase the rate of mutation or hypermutation. Thus, reactive methods inevitably give rise to a dynamic optimization problem, i.e., they model changes to the setup of the evolutionary algorithm that are not a direct result of the optimization process. For example, if we define a concrete threshold of diversity beyond which we change the rules of mutation or selection \cite{grefenstette1992genetic,ursem2002diversity}, the individual will experience ``external factors'' changing its relative fitness. Problems like these are an interesting field of research but considered outside the scope of this paper which focuses on purely static optimization problems.

Structural methods for population-based diversity attempt to construct a setup of the evolutionary algorithm that inherently favors higher-diversity results, usually without ever measuring the obtained level of diversity directly. We considered two variants of such approaches for the experiments of this paper.

\paragraph{Hypermutation.} Hypermutation (sometimes also called migration \cite{grefenstette1992genetic}) is an evolutionary operator that simply generates new individuals at random (like when constructing the initial population) and adds them to population. In early research, this was considered a dedicated method to increase the diversity of evolutionary algorithms. However, we consider the application of hypermutation to be a state-of-the-art technique for evolutionary algorithms and implemented hypermutation for all instances of evolutionary algorithms shown in this paper. This paper focuses on improvements of diversity beyond that of hypermutation, i.e., improved diversity through targeted measures instead of ``just'' increased randomness.

\paragraph{Ensembles.} Ensemble methods instantiate a number of populations at the same time. This action alone should make it more unlikely that all the so-called subpopulations converge towards the same local optimum. Additionally, these subpopulation may still interact in a limited manner. The notion of migration in this context describes the evolutionary operator that exchanges select individuals between these subpopulations. Such population structures are often called \emph{island models} and may introduce arbitrary complex rules for migration and mutual influence \cite{tomassini2005spatially}. For this paper, we considered a basic ensemble model with random migration.

\subsection{Individual-based Diversity}

Individual-based methods alter the fitness function to account for diversity. They are thus related to multi-objective evolutionary algorithms in that they construct an evolutionary process that pursues both the optimization of its objective goal function and the maximization of diversity. However, handling multi-objective evolutionary algorithms is a huge field of research, which we feel brings unneeded complexity towards a comparison of diversity measurement techniques. Thus, we only consider methods that integrate the objective goal function and the diversity measurement into a fitness function returning a scalar value.

\paragraph{Fitness Sharing.} Fitness sharing is one of the original \emph{niching} techniques \cite{mahfoud1995niching,sareni1998fitness}, first introduced in \cite{holland19751975}. It adjusts the value of the objective goal function with respect to the density of similar individuals in the population, i.e., when multiple individuals have very similar genomes, they also need to share the objective goal value achieved by these genomes. Formally, the fitness of an individual $x$ in a population $P$ is defined as \begin{equation} f(x) = \frac{g(x)}{\sum_{x' \in P} sh(x, x')}\end{equation} where the sharing factor $sh$ is given as $$sh(x, x') = \begin{cases}
    1 - (\frac{d(x, x')}{\sigma})^\alpha,& \text{if } d(x, x') < \sigma\\
    0,              & \text{otherwise}
\end{cases}$$ where $\alpha$ and $\sigma$ are parameters to the fitness sharing method. For a more in-depth explanation, please refer to \cite{sareni1998fitness}. It is important to note, however, that we also require a distance function $d$ that returns some metric of the distance $d(x, x')$ between individuals $x$ and $x'$. We will employ such a distance function directly in the paragraph on distance-based fitness and discuss the shortcomings of such a requirement there.

\paragraph{Distance-based Diversity.} Given a distance function $d: D \times D \to \mathbb{R}$ we can also directly reward individuals that stay ``far away'' from the rest of the population, thus augmenting diversity in the population. We can then simply define the fitness function as \begin{equation}f(x, P) = g(x) + \lambda * \sum_{x' \in P} \frac{d(x, x')}{|P|}\end{equation} where $\lambda$ is the weighting factor of objective goal function versus distance. Without loss of generality, we can assume that $d$ is normalized so that $0 \leq d(x, x') \leq 1$ for all $x, x' \in D$. This causes the whole term to the right of $\lambda$ to be contained in $[0; 1]$ as well, thus giving a more intuitive interpretation to the value of $\lambda$. Our experiments show that a good choice for $\lambda$  is roughly around the average objective goal value achieved by a non-diverse evolutionary algorithm. Still, we performed grid search anew every time.

The employed distance function $d$ is, of course, another parameter for this algorithm. In this paper, we mainly consider problem domains $D = \mathbb{R}^n$ for some $n \in \mathbb{N}$ where the Manhattan distance is a readily available choice of distance function. For spatial problems, the geometric distance may also be applicable in some cases. Furthermore, we also consider an instance of integer combinatory problems where Manhattan distance is meaningless but can easily substituted by Hamming distance. For a fair comparison, in this paper, we deliberately only selected problem domains where suitable distance functions can easily be given. But of course, there also exist a multitude of problem domains where the genome is given as a tree structure, or a segment of program code, or a combination of various data structures. Defining a good distance function for these domains can be a complex engineering tasks in itself, which is why we researched diversity measures that do not depend on the problem domain to such an extent.

\paragraph{Randomized Distance-based Diversity.} The distance-diverse fitness function as given above has a severe issue when applied in practical applications: The complexity of the fitness evaluation of a population is increased to $O(|P|^2)$, assuming the fitness evaluation of a single individual can be done in constant time with respect to the population size. This makes pure distance-based diversity a computationally expensive approach over the course of the evolutionary process. However, in accordance with \cite{beyer2000evolutionary} we found that evolutionary algorithms perform very robust with respect to random influences on their fitness. We can thus choose to only estimate the average distance of a single individual from the population by computing its distance to a random subset of that population. Let $S(P) \subset P$ be a random subset of the population. We can then write the distance-based fitness function as \begin{equation}f(x, P) = g(x) + \lambda * \sum_{x' \in S(P)} \frac{d(x, x')}{|S(P)|}.\end{equation} We found that selections with $|S(P)| = 5$ already performed well enough so that little difference from evaluation against the whole population could be found. Due to this fact, we employ this method of randomization whenever possible for the experiments described in this paper.

\paragraph{Inherited Fitness.} Inherited fitness allows the fitness of an individual to be influenced by the fitness of its ancestors \cite{chen2002fitness}. Alongside large parts of its genome, any individual generated via mutation or recombination thus inherits (an approximation of) its parents fitness value. Formally, we can write \begin{equation}f(x, P) = (1-\kappa) * g(x) + \kappa * h(x)\end{equation} where $$h(x) = \begin{cases} f(x') & \text{if } x \text{ is a mutation of } x',\\ \frac{f(x_1) + f(x_2)}{2} & \text{if } x \text{ is a recombination of } x_1 \text{ and } x_2,\\ 0 & \text{otherwise,}\end{cases}$$ with $\kappa \in \mathbb{R}, 0 < \kappa < 1$ being a \emph{relative} weighting factor of the inherited fitness versus the currently computed objective value.

This approach attempts to aid diversity by slowing down the process of convergence: Even when individuals with very high objective goal values are discovered, it takes them a few generations to reach their full potential fitness value. This gives individuals in other niches more time to perhaps discover competitive solution candidates. In contrast to the individual-based approaches we introduced previously, inherited fitness does not introduce additional dependencies on the problem domain but operates solely on the fitness values generated with the help of the objective goal function that is given anyway. However, we recognize that the concept of diversity induced by this model is relatively weak since the combined fitness function does not depend on other peers in the population but only on the respective individual's parents.

\paragraph{Exact Genealogical Diversity.} In an attempt to combine the benefits of distance-based diversity and inherited fitness, we introduce the notion of genealogical diversity \cite{gabor2017genealogical}. In the end, we want to achieve a functional metric of the diversity of a single individual with respect to its current population without depending on any additional domain-specific knowledge such as distance functions. Instead, we want to use the knowledge already generated by the evolutionary process to give an estimate of the diversity of single individuals. This knowledge mainly stems from the application of evolutionary operators, i.e., it contains the genealogical relations of each individual. The ulterior idea behind this approach is that we can estimate that individuals that are closely related are less likely to be diverse with respect to each other. In short, you likely are more different from your cousin twice-removed than from your child.

We annotate every individual with a map $G: D \times D \to \mathbb{R}$ containing its genealogical distance to each other individual in the population. Whenever we generate an individual, it inherits the list of its parents and updates it accordingly, i.e., assuming $x$ results from the recombination of $x_1$ and $x_2$ we can assign to $x$ $$G(x, x') := \begin{cases} 0 & \text{if } x' = x\\ r + \min\{G(x_1, x'), G(x_2, x')\} & \text{otherwise,}\end{cases}$$ where $r$ is a parameter describing the distance we value a parent-child relationship with (usually $r=1$). A similar definition can be made for children produced by mutation. Note that when individuals $x$ and $x'$ are completely unrelated, for example when one of them was newly generated by hypermutation, we assign some maximum value $t$ which we also divide the results of $G$ by in order to achieve normalization again. The resulting fitness function then looks pretty standard as \begin{equation}f(x, P) = g(x) + \lambda * \frac{\sum_{x' \in S(P)}G(x, x')}{|S(P)| * t}.\end{equation}

\paragraph{Genealogical Fitness.} When directly applying exact genealogical fitness as described above, we again run into a complexity issue: We need to save the information of the whole tree of genealogical relations produced by the evolutionary process. Even if we limit ourselves to the distances between individuals still present in the population, we end up with a spatial complexity $O(|P|^2)$. For the populations used for the experiments, this was manageable and we thus performed these experiments for exact genealogical fitness as well. However, we still want an approach that scales well even with much larger populations. To this end, we were inspired by the way researches in biology determine the relatedness between singular individuals: They match their genomes. Compared to biological systems artificial evolution usually features much smaller genomes. Furthermore, large part of biological genomes are actually not subjected to selection pressure and can thus record patterns (and by extent genealogical relationships) without bias. We try to mimic these properties for our final approach towards diversity-aware evolutionary algorithms.

%

In order to efficiently approximate the genetic relation between two given individuals (without keeping a complete history of the whole evolutionary process), we assign every individual a bitstring $b = b_0, ..., b_{\tau-1}$ of length $\tau$. Note that $\tau$ is the only hyperparameter introduced by genealogical diversity (and we noticed to be very robust with respect to different choices of $\tau$). Every time an operator like mutation or crossover is applied to some individuals, we apply the respective operator to their assigned bitstrings. Since bitstrings are a classic among the representations used in genetic algorithms, most operators (even when designed for other data structures) have an immediate counterpart defined on $\{0, 1\}^{*}$, which is the alphabet of bitstrings. Whenever an individual is newly created, it is assigned a random bitstring. When we choose $\tau$ large enough, these random bitstrings will feature a relatively high Hamming distance to each other. Throughout the process of evolution, we will interpret a high Hamming distance between two individuals as a sign of non-relatedness. Note that the use of the Hamming distance here does not depend on the problem domain but only on our choice to use bitstrings to augment the problem-specific genomes. More specifically, genealogical diversity is defined via a genealogical distance function $d: P \times P \to \mathbb{N}$, which can be written as $$d(x, x') = \frac{1}{\tau} * \sum_{i=0}^{\tau - 1} \begin{cases}
0 & \textit{if } b_i = b_i'\\
1 & \textit{otherwise}
\end{cases}$$ where $b_0, ..., b_{\tau-1}$ is the bitstring assigned to $x$ and $b_0', ..., b_{\tau-1}'$ is the bitstring assigned to $x'$. We then again can simply write \begin{equation}f(x, P) = g(x) + \lambda * \sum_{x' \in P}\frac{d(x, x')}{|S(P)|}\end{equation} for the genealogically diverse fitness function. During the course of evolution, the bitstrings are not subject to selection but are subjected to the variational evolutionary operators. Thus, the bitstring can record the degree of relatedness between individuals, albeit in a highly probabilistic manner. Still, evolutionary algorithms show robustness with respect to the added noise in selection and comparing a few bitstrings to estimate diversity is highly efficient in both time and space complexity (with $\tau \approx log(|P|)$ being a good setting from an experimenter's experience).

%
%

\section{Experimental Results}
\label{sec:results}

To analyze the effectiveness of the various forms of diversity-respecting evolutionary algorithms described in this paper, we have considered various settings. In Subsection~\ref{subsec:schwefel} we first describe a common benchmark problem often used to evaluate evolutionary algorithms \cite{benchmarks}. Subsection~\ref{subsec:robot} describes the setup and evaluation of a navigational problem written as a real vector optimization. Finally, Subsection~\ref{subsec:factory} considers an integer combinatory problem.

\subsection{The Schwefel Problem}
\label{subsec:schwefel}

\begin{figure}[t]
  \centering
  \includegraphics[width = 0.4\textwidth]{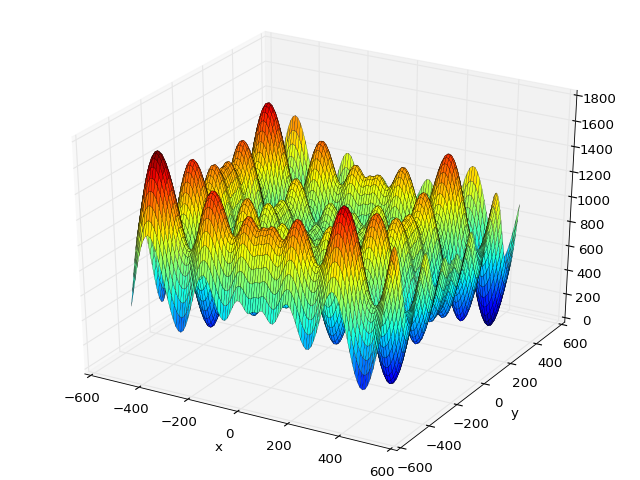}
  \caption{Illustration of the Schwefel problem in two dimensions. Taken from \cite{benchmarks}.}
  \label{fig:schwefel-problem}
\end{figure}

The Schwefel problem has often been used as a benchmark problem for evolutionary algorithms \cite{gordon1993serial,deb1999multi}. For this experiment, we used the implementation given by the DEAP library \cite{fortin2012deap,benchmarks}. The Schwefel problem is parametrized on the dimension of the search space (we simply write $|D|$) and given as the function \begin{equation}g(x) = 418.9828872724339 * |D| - \sum_{i=1}^{|D|} x_i * \sin(\sqrt{|x_i|})\end{equation} with the optimal solution to the minimization problem being $(0)^{|D|}$. Figure~\ref{fig:schwefel-problem} provides a visualization of the solution landscape. 

\begin{figure}[t]
  \centering
  \includegraphics[width = 0.48\textwidth]{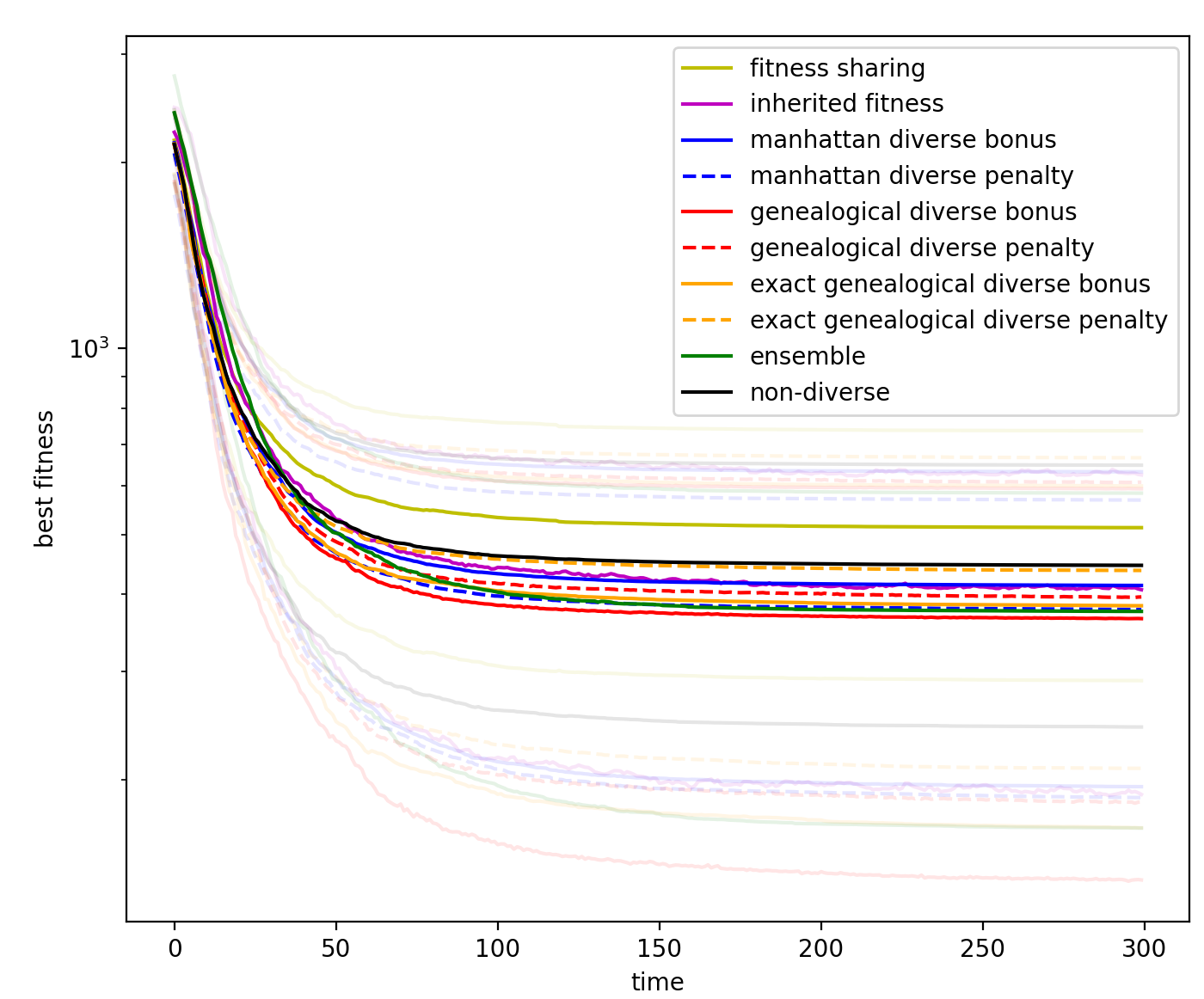}
  \caption{Evaluation results for the Schwefel problem. For each generation, we plot the current population's  best objective value on a log scale. Averaged over 100 independent runs. Semi-transparent lines show plus/minus one standard deviation.}
  \label{fig:schwefel-results}
\end{figure}

\paragraph{Setup.} We show an evaluation of the Schwefel problem with $|D| = 8$. For all evolutionary algorithms, we used a total population size of $30$ and ran evolution for $300$ generations. For all experiments, we used a single-spot mutation operator and uniform recombination. We consistently chose relatively high values for variational parameters by opting for a mutation rate of $0.1$, a recombination rate of $0.3$, and a hypermutation rate of $0.1$. We did so to have all algorithms benefit from diversity through increased randomness in the evolutionary process and thus evaluate their ability to produce diversity beyond adding random noise.
For fitness sharing, we set $\alpha = 2$ and $\sigma$ to the maximum value, so that it spans the whole problem domain. Inherited fitness used $\kappa = 0.2$. Genealogical diverse algorithms used a bitstring size of $\tau = 16$. The ensemble approach split the population into $3$ subpopulations with a random migration rate of $0.1$.
For all weighted diversity mechanisms (those featuring a weighting factor $\lambda$ in their fitness function) we chose $\lambda = 200$ for the Schwefel experiment. All parameters were approximated for best performance via manual grid search.
For all experiments performed for this paper, we tested two variants of the weighted diversity mechanisms. In one case, we used the diversity term as described throughout this paper, which means it adds a bonus value in the direction of the optimization process. We also evaluated variants that (instead of rewarding high diversity values) penalize low diversity values by moving the respective individuals away from the optimization goal. Unsurprisingly, no real difference was observed in this regard.

\paragraph{Results.} Due to the high amount of local optima it proved difficult for all tested algorithms to find the global optimum. The results in Figure~\ref{fig:schwefel-results} show that none of the algorithms reached the minimum in the given time. While all algorithms perform extremely similar, the genealogical variant can be seen at the lowest point, although without too significant of a difference.

We ran this experiment for other benchmark problems contained in DEAP, notably H1, Schaffer and Rosenbrock \cite{deb1999multi,benchmarks}. But the difference between various algorithms was even smaller for these experiments, which is why we left out the respective plots. All software and results are available online.\footnote{\url{gitlab.lrz.de/thomasgabor/gecco-evolib}}

\subsection{The Pathfinding Problem}
\label{subsec:robot}

\begin{figure}[t]
  \centering
  \includegraphics[width = 0.4\textwidth]{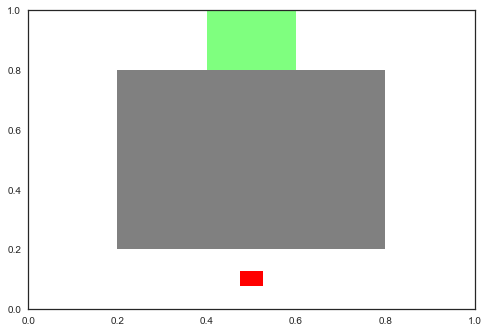}
  \vspace*{-3mm}
  \caption{Illustration of the Pathfinding problem. A robot (red) gets rewarded for each of the 10 time steps of its life that it spends in the target area (green). It thus needs to reach the goal quickly using steps of size 0.3 in each dimension.}
  \label{fig:pathfinding-problem}
\end{figure}

\begin{figure*}[t]
  \centering
  \includegraphics[width = 0.94\textwidth]{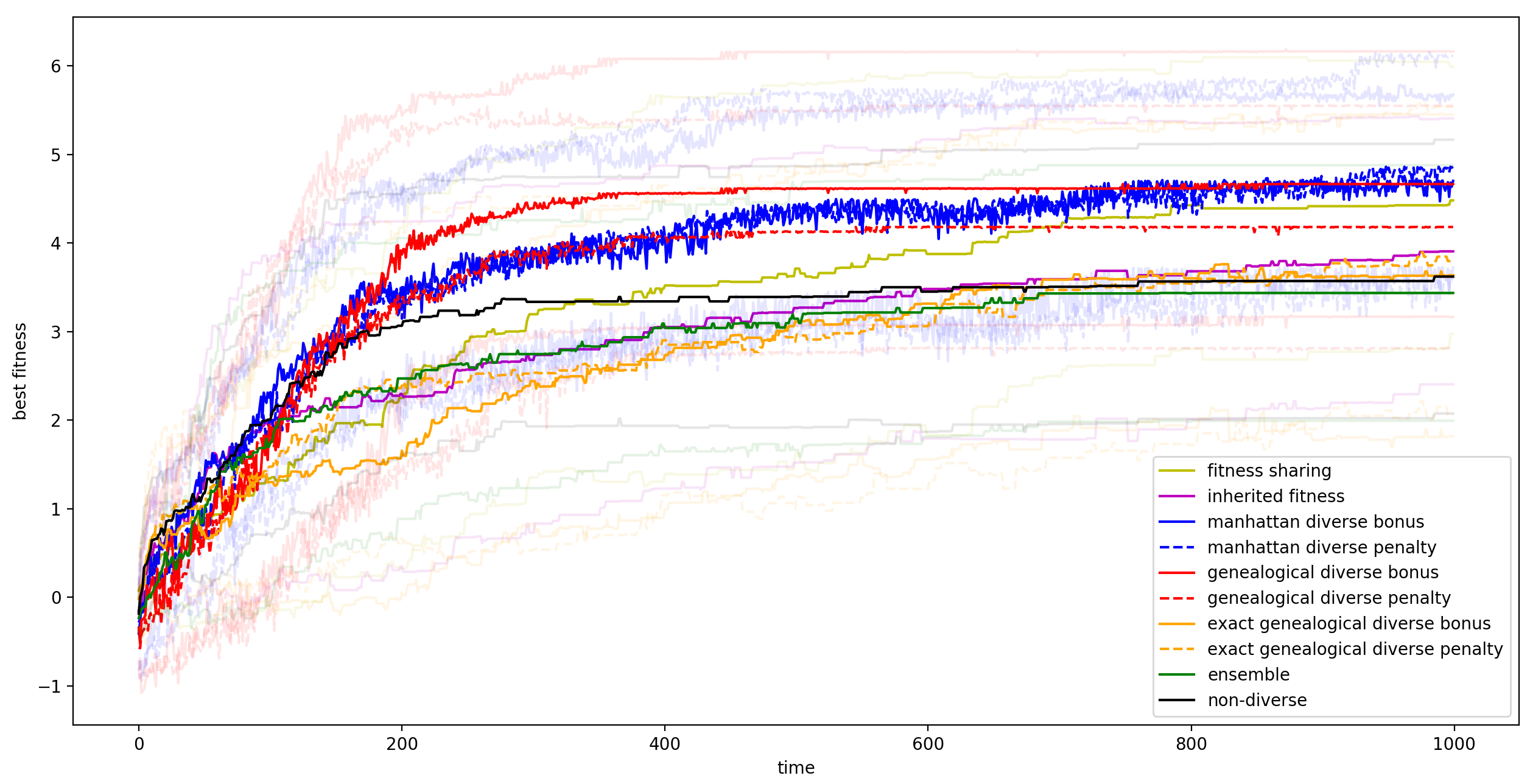}
  \vspace*{-3mm}
  \caption{Evaluation results for the Pathfinding problem. For each generation, we plot the current population's  best objective value. Averaged over 20 independent runs. Semi-transparent lines show plus/minus one standard deviation.}
  \label{fig:pathfinding-results}
\end{figure*}

Given a room of dimensions $1 \times 1$, we imagine a robot standing at position $(0.5, 0.1)$. It needs to reach a goal area at the opposite side of the room, given by the square of side length $0.2$ centered around the point $(0.5, 0.9)$. However, the room features a huge obstacle between those points and thus the robot needs to decide on a way around it. The agent can move by performing an action $a \in \{(\delta{x}, \delta{y}) | -0.3 < \delta{x} < 0.3, -0.3 < \delta{y} < 0.3\}$. The robot needs to develop a plan consisting of $10$ such actions that will get it to the goal area. Once it has reached that area, it gets rewarded for staying there as long as possible. This setup is illustrated in Figure~\ref{fig:pathfinding-problem}.

\paragraph{Setup.} The Pathfinding problem has a dimension of $|D| = 20$ with $D = (\delta{x_1}, \delta{y_1}, \delta{x_2}, \delta{y_2}, ..., \delta{x_{10}}, \delta{y_{10}})$. The robot earns a reward of $1$ for each time step spent within the goal area and receives a penalty of $-0.1$ for each attempt to perform an illegal step, i.e., a step ending up outside the room or inside the obstacle. Illegal steps are disregarded entirely (so the robot does not move \emph{up to} the wall when attempting to step beyond it).

For the experiment, we used a population size of $100$ individuals for all evolutionary algorithms. We allowed them to run for $1000$ generations. Again, we used $\alpha = 2$ and $\sigma = \max$ for fitness sharing, split the population into $3$ subpopulations for the ensemble method (yielding $34, 33, 33$ for the subpopulation sizes), and set $\tau = 16$ for the genealogical algorithm. For all weighted diversity mechanisms we used $\lambda = 12$ this time, putting a high stress on diversity.

\paragraph{Results.} For the Pathfinding problem, favoring diversity pays off in the optimization result. The results are shown in Figure~\ref{fig:pathfinding-results}. Distance-based diversity in the form of Manhattan diversity performed best. Genealogical diversity is a close second, however, achieving similar levels of results without a domain-specific distance function. On third place, fitness sharing too reaches similar results but is computationally more expensive. Using exact genealogical diversity, which was the original motivation for genealogical diversity, seems to have little effect on the result. We argue that parameters like the relative weight of recombination and mutation relations require further tweaking towards the problem-specific requirements. We find, however, that this defeats the original purpose of employing inheritance-based diversity measures in the first place. We thus focus on the results of the bitstring-augmented genealogical diversity instead. On a surprising note, both inherited fitness and the ensemble model perform worse than the standard evolutionary algorithm. Both may show a slowing effect on the evolutionary process. This means that for this problem, different random initialization of subpopulation most likely plays no role in enhancing diversity as even remotely competitive solution candidates are only found later on. It thus seems that these methods may in fact tackle different classes of problems.

\subsection{The Routing Problem}
\label{subsec:factory}

\begin{figure}[t]
  \centering
  \includegraphics[width = 0.4\textwidth]{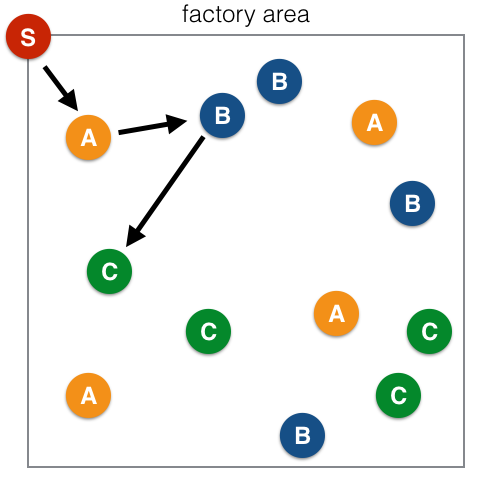}
  \vspace*{-5mm}
  \caption{Illustration of the Routing problem. A robot \emph{S} (red) needs to drive to workstations of types A, B, C in order. For each workstation type, it can choose from several workstations inside the factory. We need to find the routing path that minimizes the travelled distance.}
  \label{fig:routing-problem}
\end{figure}

For the last experiment, we wanted to opt for a discrete combinatorial problem in contrast to the continuous optimization of real-valued vectors performed so far. We again imagined a robot as it may work in a smart factory in the near future. This time, the robot already knows how to best travel to any given target, maybe involving various means of transport like forming convoys of robots or using conveyor belts installed in the factory. The robot is given the task to travel to various workstations that exist inside the factory in order to process a specific item it is carrying around. This item needs a certain amount of tasks to be performed in order to be produced. For each of these tasks, there are $5$ dedicated workstations scattered throughout the factory. Figure~\ref{fig:routing-problem} shows a smaller instance of that setup.

\paragraph{Setup.} For this experiment, we choose a setting with $12$ different tasks, resulting in a factory with $60$ workstations. Accordingly, solution candidates are of the type $\{1, 2, ..., 5\}^{12}$. The genome $(2, 4, ...)$, e.g., ist interpreted as ``go to the workstation of type A with the number 2; go to the workstation of type B with the number 4; ...'' so that no type mismatch can ever happen. To mimic various means of transport, we randomized the distance between each of these workstations individually within a range of $[0, 100] \subset \mathbb{R}$. Note that this (most likely) gives rise to a non-euclidean space the robot is navigating, making the problem as difficult as finding the shortest weighted path in an arbitrary tree.

For the parameters of evolution, we used a population of $50$ and ran each algorithm for $100$ generations. Fitness sharing kept $\alpha = 2$ and $\sigma = \max$ but employed the Hamming distance instead of Manhattan distance to compute the niching radii. Inherited and genealogical fitness kept $\kappa = 0.2$ and $\tau = 16$, respectively. Ensemble model again used $3$ subpopulations. For distance-based diversity, the Manhattan distance is no longer applicable since there is no associated meaning with the numbering of the workstations of a specific type. In this case, we can simply use Hamming distance instead. But it shows that for more complicated genome types, additional engineering effort may be necessary here. For all weighted diversity mechanisms we set $\lambda = 250$.

\begin{figure}[t]
  \centering
  \includegraphics[width = 0.48\textwidth]{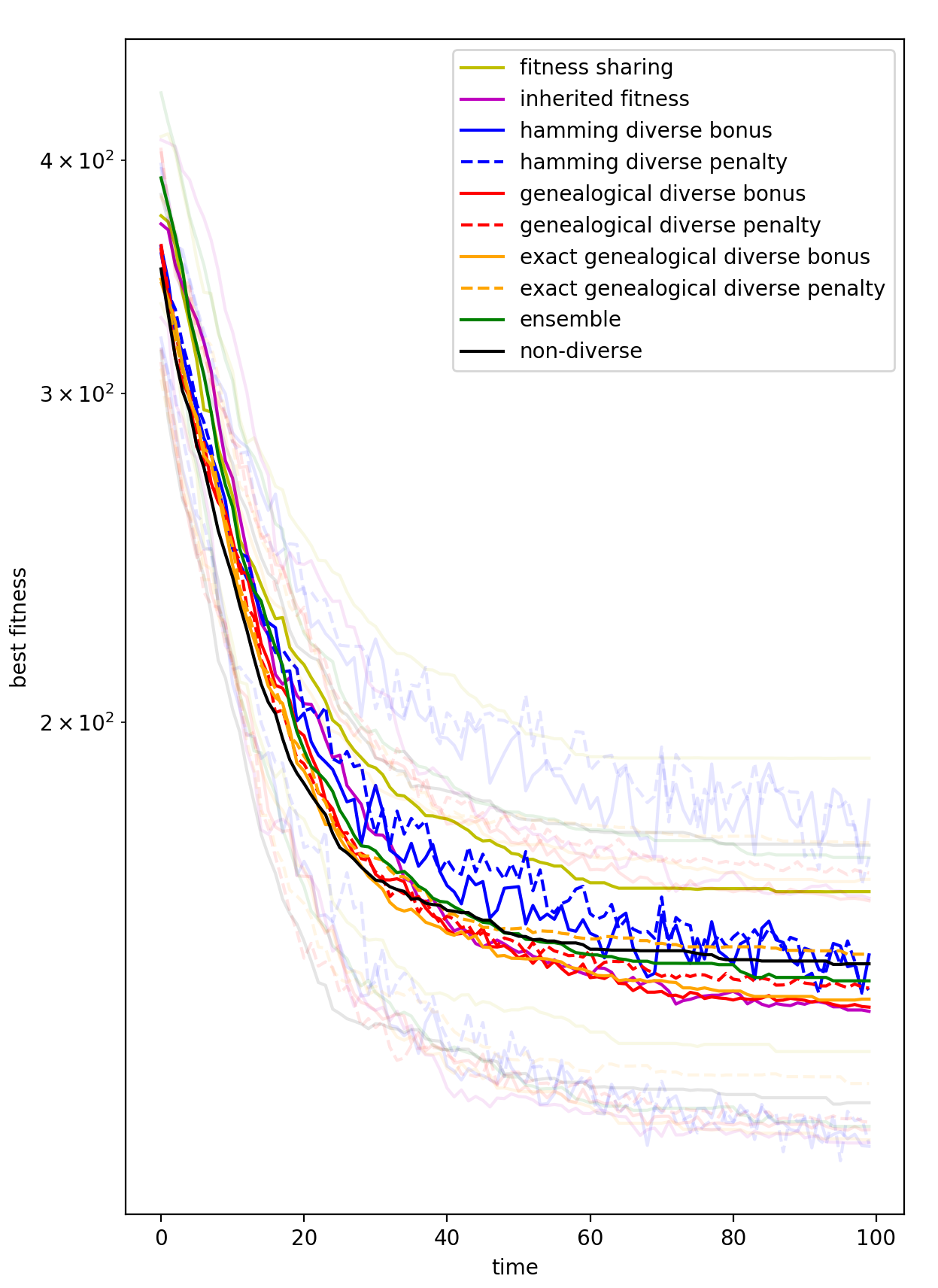}
  \vspace*{-5mm}
  \caption{Evaluation results for the Factory Routing problem. For each generation, we plot the current population's best objective value on a log scale. Averaged over 100 independent runs. Semi-transparent lines show plus/minus one standard deviation.}
  \label{fig:routing-results}
\end{figure}

\paragraph{Results.} The results of the factory routing experiment are depicted in Figure~\ref{fig:routing-results}. We observe that fitness sharing seems to not perform as well using the Hamming distance function. The exact genealogical diversity approach again seems to make not that much of a difference. In this setting, so does the ensemble approach. However, the other three means to establish diversity do manage to achieve slightly better results. They all perform on a comparable level in the best cases, with Manhattan-based diversity being subject to more fluctuation than inherited or genealogical fitness.

\section{Conclusion}
\label{sec:conclusion}

While diversity has been known to be an important factor for the analysis of evolutionary algorithms, we focused on the explicit integration of diversity into a single-objective fitness function, which is a method not yet fully explored. Connections of this approach to standard multi-objective evolutionary algorithms and respective approaches that add diversity as an additional full-fledged objective are still to be researched. Furthermore, we focused on the issue of global optimization, i.e., we evaluated the tested algorithms for their ability to better approximate the global optimum only instead of, e.g., achieving a better coverage of various local optima \cite{vassiliades2017comparing}. We tested the approach of explicit weighted integration against common diversity techniques like fitness sharing, ensemble evolutionary algorithms or inherited fitness.

While domain-specific distance functions have been evaluated to be the most successful in several examples, we also aimed to provide means of measuring diversity that can more simply be plugged into existing algorithms (and libraries) without requiring as much domain-specific adjustments. For this purpose, we motivated and introduced the novel approach of genealogical diversity for a full evaluation. Inspired by nature, this approach augments the genomes by data structures not subjected to selection bias. We can then trace relatedness between individuals by analyzing the matches in these additional genes. The exact requirements on the size of these augmentations are still up to future research.

After all, it seems that several classes of optimization problems may be discerned here. Several benchmark problems have shown to be hardly affected by the additional stress on diversity while example problems motivated by industrial scenarios with the need to apply optimization techniques in practical applications benefit to a relatively large extent from the explicit treatment of diversity. How and when this is the case needs further research. Further reduction of hyperparameters seems to be an important step for a fair and broad evaluation of a multitude of approaches. We already suggested a rule of thumb for the setting fo diversity weight $\lambda$ but an extensive study on this matter is still missing. It may be possible to automatically set $\lambda$ to appropriate values just as the mutation rate can usually be left to be determined by the algorithm itself \cite{eiben2003introduction}.

At least in theory, inheritance-based diversity estimation methods need not be limited to the past. In the end, the ulterior motive to employ diversity is to favor individuals that will \emph{eventually} give rise to the best solution candidates. Obviously, this cannot be accurately predicted without actually executing the whole evolutionary process. This usually is physically impossible for all but small problem instances. But perhaps, this property can be approximated. Diversity should then favor individuals that \emph{cover} a lot of good options after the application of the evolutionary operators over individuals that \emph{are} a good option. We suggest this as an important direction for future research.


\bibliographystyle{ACM-Reference-Format}
\bibliography{references}

\end{document}